\documentclass[journal,twoside,web]{ieeecolor}
\usepackage{generic}
\usepackage{cite}
\usepackage{amsmath,amssymb,amsfonts}
\usepackage{multirow}
\usepackage{algorithmic}
\usepackage{graphicx}
\usepackage{textcomp}
\usepackage{lineno,hyperref}
\def\BibTeX{{\rm B\kern-.05em{\sc i\kern-.025em b}\kern-.08em
    T\kern-.1667em\lower.7ex\hbox{E}\kern-.125emX}}
\begin{document}
\title{Deep Learning Methods and Applications for Region of Interest Detection in Dermoscopic Images}
\author{Manu Goyal, Moi Hoon Yap, and Saeed Hassanpour, \IEEEmembership{Member, IEEE}
\thanks{This research was supported in part by National Institutes of Health grants R01LM012837 and R01CA249758. We submitted this work for review in IEEE Jorunal of Biomedical and Health Informatics}
\thanks{Manu Goyal is with the Department of Biomedical Data Science, Dartmouth College, Hanover, NH, USA. He is now with the Department of Radiology, UTSouthwestern Medical Center, Dallas, TX 75390 USA  (e-mail: manu.goyal@dartmouth.edu; manu.goyal@utsouthwestern.edu). }
\thanks{Moi Hoon yap is with Department of Computing and Mathematics, Manchester Metropolitan University, Manchester, M1 5GD, UK. (e-mail: m.yap@mmu.ac.uk).}
\thanks{Saeed Hassanpour is with Departments of Biomedical Data Science, Computer Science, and Epidemiology, Dartmouth College, Hanover, NH, USA (e-mail: saeed.hassanpour@darmouth.edu).}}

\maketitle

\begin{abstract}
Rapid growth in the development of medical imaging analysis technology has been propelled by the great interest in improving computer-aided diagnosis and detection (CAD) systems for three popular image visualization tasks: classification, segmentation, and Region of Interest (ROI) detection. However, a limited number of datasets with ground truth annotations are available for developing segmentation and ROI detection of lesions, as expert annotations are laborious and expensive. Detecting the ROI is vital to locate lesions accurately. In this paper, we propose the use of two deep object detection meta-architectures (Faster R-CNN Inception-V2 and SSD Inception-V2) to develop robust ROI detection of skin lesions in dermoscopic datasets (2017 ISIC Challenge, PH2, and HAM10000), and compared the performance with state-of-the-art segmentation algorithm (DeeplabV3+). To further demonstrate the potential of our work, we built a smartphone application for real-time automated detection of skin lesions based on this methodology. In addition, we developed an automated natural data-augmentation method from ROI detection to produce augmented copies of dermoscopic images, as a pre-processing step in the segmentation of skin lesions to further improve the performance of the current state-of-the-art deep learning algorithm.  Our proposed ROI detection has the potential to more appropriately streamline dermatology referrals and reduce unnecessary biopsies in the diagnosis of skin cancer. 
\end{abstract}



\section{Introduction}

According to the Skin Cancer Foundation \cite{SkinCancerFoundation}, the incidence of new skin cancer cases is higher than the combined number of new cases for breast, lung, colon, and prostate cancer over the past three decades. In current medical practice, dermatologists primarily examine patients by visual inspection with a dermatoscope to determine the condition of skin lesions. Relying on clinicians' self-vigilance and vision to examine skin lesions risks patients' lives and survival rates, as it is challenging to identify the type of skin lesion with the naked eye alone. Dermoscopy is non-invasive imaging that allows visualization of the skin surface by using a light magnifying device and immersion fluid \cite{pellacani2002comparison}. It is one of the most widely used imaging techniques in dermatology, and it has increased the rate of accurate diagnosis of skin disease \cite{mayer1997systematic}.

The previous state-of-the-art computer-aided decision support tools for dermoscopic images of skin lesions are based on image processing and traditional machine learning techniques, which are composed of multi-stages that include image pre-processing, image segmentation \cite{celebi2009lesion}, feature extraction \cite{barata2018survey}, and classification \cite{celebi2019dermoscopy, 8936444}. 
Since the introduction of the largest public skin lesion dataset as part of the annual International Skin Imaging Collaboration (ISIC) Challenge, end-to-end deep learning algorithms have gained wide popularity for skin lesion classification and segmentation tasks \cite{codella2018skin}. The skin segmentation challenge consists of segmentation of the lesion's boundary, irrespective of classes of skin lesions. In one of the studies, however, Goyal et al. used deep segmentation networks (fully convolutional networks) for the multi-class diagnosis of skin lesions \cite{goyal2017multi}. Skin lesion segmentation is an essential tool to support dermatologists, and it can provide insight about varying characteristics and attributes of skin lesions, such as the Asymmetry, Border, Color, and Diameter (ABCD) rule \cite{nachbar1994abcd}. 

Region of Interest (ROI) detection has proved to be important in medical image analysis  \cite{goyal2018robust, criminisi2013regression, yap2005region}, where it is defined as a bounding box circumscribing the lesion. ROI detection by traditional machine learning methods relies on a sliding window approach that uses classifiers trained on skin lesion patches \cite{celebi2009lesion}, and it is inefficient compared to end-to-end deep learning algorithms. Han et al. proposed the use of region-based CNN for keratinocytic skin cancer detection on the clinical photographs with high sensitivity \cite{han2020keratinocytic}. 

This paper focuses on automatic detection of the ROI in dermoscopic images. We performed this study on the following datasets: 2017 ISIC Challenge  \cite{codella2017skin}; HAM10000 (a part of the 2018 ISIC Challenge dataset) \cite{tschandl2018ham10000}; and PH2 \cite{mendoncca2013ph}, which consists of different types of skin lesions, such as naevus, melanoma, seborrhoeic keratoses, and other skin lesions. Skin lesions have significant intra-class variations in terms of color, size, place, and appearance for each class and high inter-class visual similarities \cite{goyal2017multi}. Also, researchers augmented the data by using image manipulation techniques to overcome the lack of training data in deep learning. Tschandl et al. \cite{tschandl2018ham10000} used a camera with different magnifications and angles to capture the same skin lesion to produce augmented copies known as natural data-augmentation for the HAM10000 dataset. However, it means doing laborious processing and having redundant data for the same skin lesion. With the aid of our ROI detection process, we develop a natural data-augmentation method to produce augmented copies of images for skin lesion datasets. Automated ROI detection for skin lesions has great potential to improve both the quality of skin lesion datasets and the accuracy of lesion detection, and simplify the laborious task of manual annotation. The key contributions of this paper are as follows:


\begin{enumerate}
	\item We propose the use of deep learning algorithms for ROI detection on dermoscopic images of the 2017 ISIC Challenge dataset. Then, we test the robustness of trained models on other completely unseen publicly available datasets, i.e., the PH2 and HAM10000 datasets.
	\item We demonstrate the practical application of this work by developing an android camera application, which utilizes these models in mobile devices for real-time ROI detection of skin lesions. Furthermore, real-time ROI detection can be used to capture a highly standardized skin lesion dataset using a smartphone camera.
	\item We demonstrate the use of the ROI detection algorithm to produce automated natural data-augmentation. We test the performance of deep learning algorithms with the use of natural data-augmentation as a pre-processing step for lesion segmentation.
\end{enumerate}

\section{Methodology}
This section presents a brief description of skin lesion datasets (2017 ISIC Challenge, PH2, and HAM10000) used in this work; the conversion of the segmentation ground truths to bounding box ground truths; and our ROI detection deep learning algorithm. Finally, we describe the performance evaluation using popular ROI detection evaluation metrics.

\subsection{Publicly Available Skin-Lesion Datasets}
For this experiment, we utilized the 2017 ISIC Challenge dataset, which consists of a training dataset (2,000 images), a validation dataset (150 images), and a testing dataset (600 images) for ROI detection of skin lesions. This dataset includes dermoscopic images of three types of skin lesions: nevi, melanoma, and seborrhoeic keratosis \cite{codella2017skin}. 

\begin{figure}
	\centering
	\includegraphics[width=0.5\textwidth]{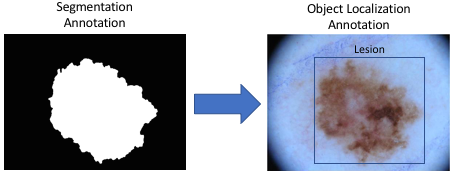}
	\caption{Conversion of segmentation ground truths to detection ground truths in ISIC Challenge 2017 dataset.}
	\label{fig:GT}
\end{figure}

The PH2 and HAM10000 datasets are utilized as external testing datasets further to validate the performance of our ROI detection algorithm. The PH2 dataset consists of 200 dermoscopic images of two types of skin lesions: nevi (160 images) and melanoma (40 images) \cite{mendoncca2013ph}. Since the ground truths for the challenges only included the segmentation mask, to produce the ground truth for ROI, we circumscribed a rectangular bounding box on the segmentation mask (Fig. \ref{fig:GT}). The HAM10000 dataset is a recent publicly available skin lesion dataset that consists of 11,788 dermoscopic images of different skin lesions collected from multiple sources around the world \cite{tschandl2018ham10000}. The HAM10000 dataset is a subset of the 2018 ISIC Classification Challenge  dataset consists of seven different types of dermoscopic skin lesions: nevi, melanoma, seborrheic keratosis, actinic keratosis, basal cell carcinoma, dermatofibroma, and vascular lesion. Since nevi, melanoma, and seborrheic keratosis are already available in the 2017 ISIC Challenge dataset and the PH2 dataset, we randomly selected a subset of images from the HAM10000 dataset by picking 50 images from each of four categories of skin lesions: actinic keratosis, basal cell carcinoma, dermatofibroma, and vascular lesion.  Since these types of lesions are not available in the training dataset, we used this subset of images from the HAM10000 dataset to evaluate the generalizability of CNNs for ROI detection of other types of skin lesions. There is no ground truth annotations are available for developing ROI detection of lesions in this dataset.  Hence we manually annotated the ground truths of ROI of skin lesions for this subset of HAM10000 dataset. To build these datasets, dermoscopic images were captured with high-resolution cameras. For this experiment, we resized the dermoscopic images to 500$\times$375 to reduce the training time and improve the inference speed of ROI detection algorithms; ultimately, the size of the images was similar to the images in the 2012 PASCAL VOC dataset \cite{8936444, everingham2010pascal}. The number of dermoscopic images used from above mentioned datasets for this work is described in Table \ref{DatasetSplit}.

\begin{table*}[]
	\centering
	\addtolength{\tabcolsep}{4pt}
	\renewcommand{\arraystretch}{2}
	\caption{Summary of dermoscopic images used for training, validation and testing sets where \textit{mel} is Melanoma, \textit{nev} is Nevi, \textit{sk} is Sebohorriec Keratosis, \textit{ak} is Actinic Keratosis, \textit{bk} is Benign Keratosis, \textit{dmf} is Dermatofibroma, \textit{bcc} is Basel Cell Carcinoma, \textit{vasc} is Vascular Lesion}
	\label{DatasetSplit}
	\scalebox{0.75}{
		\begin{tabular}{llllllllll}\hline
			Dataset Split                 & Dataset Name             & \textit{mel} & \textit{nev}  & \textit{sk}  & \textit{bcc} & \textit{ak} & \textit{dmf} & \textit{vasc} & Total \\\hline \hline
			Training Set                  & ISIC-2017 Training Set   & 374 & 1372 & 254 & 0   & 0  & 0   & 0    & 2000  \\ \hline
			Validation Set                & ISIC-2017 Validation Set & 30  & 78   & 42  & 0   & 0  & 0   & 0    & 150   \\ \hline
			\multirow{3}{*}{Testing Sets} & ISIC-2017 Validation Set & 117 & 393  & 90  & 0   & 0  & 0   & 0    & 600   \\ 
			& PH2 dataset              & 40  & 160  & 0   & 0   & 0  & 0   & 0    & 200   \\ 
			& HAM10000 dataset        & 0   & 0    & 0   & 50  & 50 & 50  & 50   & 200  \\ \hline
	\end{tabular}}
\end{table*}

\subsection*{Data Availability}
All the datasets (\href{https://challenge.kitware.com/#challenge/583f126bcad3a51cc66c8d9a}{ISIC-2017} \cite{codella2017skin}, \href{https://www.dropbox.com/s/k88qukc20ljnbuo/PH2Dataset.rar}{PH2} \cite{mendoncca2013ph}, and \href{https://www.isic-archive.com/#!/topWithHeader/onlyHeaderTop/gallery}{HAM10000} \cite{tschandl2018ham10000}) that are used in this study are freely available to download for research purposes.

\subsection*{IRB Approval}
Institutional review board (IRB) approval was not required for this study because all of the utilized datasets were publicaly available online.

\subsection{Deep Learning Methods for ROI Detection}

The Faster R-CNN \cite{ren2015faster} is a region proposal network that uses three steps to provide faster and more accurate box proposals and classifies them as an object or background. First, a convolutional neural network (CNN) is used to extract the convolutional features from the input image. Second, box proposals of different aspects are generated based on the features. Finally, box proposals are fed to CNN for classification and regression. The classification task predicts whether the box proposals contain an object or not, and regression further improves the location of box proposals.  In general, the object detection network consists of three stages to produce ROI detection \cite{ren2015faster}. These stages are briefly explained below.


\subsubsection{CNN as Feature Extractor}
In the first step, the input image goes through a convolution neural network that outputs a set of convolutional feature maps on the last convolutional layer (Supplementary Fig. 1). We used Inception-V2, which is a lightweight CNN with new features, such as improvement normalization, and factorizing the large convolution to smaller convolutions to reduce computations \cite{szegedy2016rethinking}. The other added features include using the depth-wise convolutional layers rather than typical convolutional layers to improve the processing time further, and a batch normalization layer, which can decrease internal covariate shift and combat the gradient vanishing problem to improve convergence during training \cite{szegedy2016rethinking}.

\subsubsection{Generation of proposals}
In the second stage, a sliding window of 3$\times$3 is run spatially on these feature maps. For each location, k (k=9) proposals are generated with three aspect ratios (1:1, 1:2, 2:1) and scale sizes (128$\times$128, 256$\times$256, 512$\times$512) in the original image (Supplementary Fig. 2). Then, the Intersection of Union (IoU) metric of each proposal is compared with the ground truth (GT) of that image to select the proposal for the next stage.

\subsubsection{Proposals Classification and Regression}
The selected proposals from the second stage are classified as an object or background by the CNN and are refined further to get the final ROI detections with the help of regression (Supplementary Fig. 3).

Single Shot Multibox Detector (SSD) \cite{liu2016ssd} is a meta-architecture used for object detection, which uses a single-stage convolutional neural network to predict classes directly and anchors offsets without the need of a third stage, as used in Faster R-CNN \cite{ren2015faster}. The SSD meta-architecture can produce anchors much faster than other object detection networks, and that makes it more suitable for the mobile platforms that have more limited resources than computers (Supplementary Fig. 4).

As there are no existing ROI detection methods by CNN for dermoscopic skin lesion research is available, we compared our results with the skin segmentation algorithm. We used DeepLabV3+  \cite{chen2018deeplab}, which is one of the best semantic segmentation networks achieving the test set performance of 89.0\% and 82.1\% in PASCAL VOC 2012 and Cityscapes datasets respectively to train on the ISIC-2017 segmentation dataset \cite{8936444}. The architecture of DeeplabV3+ (Supplementary Fig. 5).


We used the TensorFlow object detection API \cite{huang2016speed}, which provides an open-source framework to design and build various object detection models. For this work, we propose the use of Inception-V2 as a base network for feature extraction and classification of anchor boxes, with all object detection meta-architectures as mentioned above for the ROI lesion detection \cite{szegedy2016rethinking}. Inception-V2 provides a balanced trade-off between speed and accuracy that is appropriate for the implementation of a smartphone application using these algorithms \cite{huang2016speed}.


\subsection{Quantitative Performance Measures for Lesion Detection}

All performance metrics are calculated with``overlap criterion" as an \textit{intersection over union} (\textit{IoU}) of the detected lesion and ground truth. A \textit{True Positive} (\textit{TP}) is when the \textit{IoU}~\textgreater~0.5. A \textit{False Positive} (\textit{FP}) is a detected ROI with $\textit{IoU}\le0.5$ and the duplicate bounding boxes. A \textit{False Negative} (\textit{FN}) is when there is no ROI detected by the algorithm.

We evaluate the performance of the proposed methods using three metrics, i.e. \textit{Precision}, \textit{Recall} and \textit{Mean IoU}. \textit{Precision} is calculated by number of \textit{TP} divided by the sum of number of \textit{TP} and \textit{FP}. \textit{Recall} is the number of \textit{TP} divided by sum of number of \textit{TP} and \textit{FN}. Intersection over Union (IoU), also known as the Jaccard index, is a popular evaluation metric used to measure the accuracy of a detection algorithm. 

\begin{equation}
	IoU=\frac{|P \cap G|}{|P \cup G|}  
\end{equation}

Where P and G are the prediction and ground truth bounding boxes, respectively, we report the \textit{Mean IoU}, which is the average of the overlap percentage of the \textit{TP} cases (detected lesions).

\begin{figure}[h]
	\centering
	\begin{tabular}{cccc}
		\includegraphics[width=2.5cm,height=2.5cm]{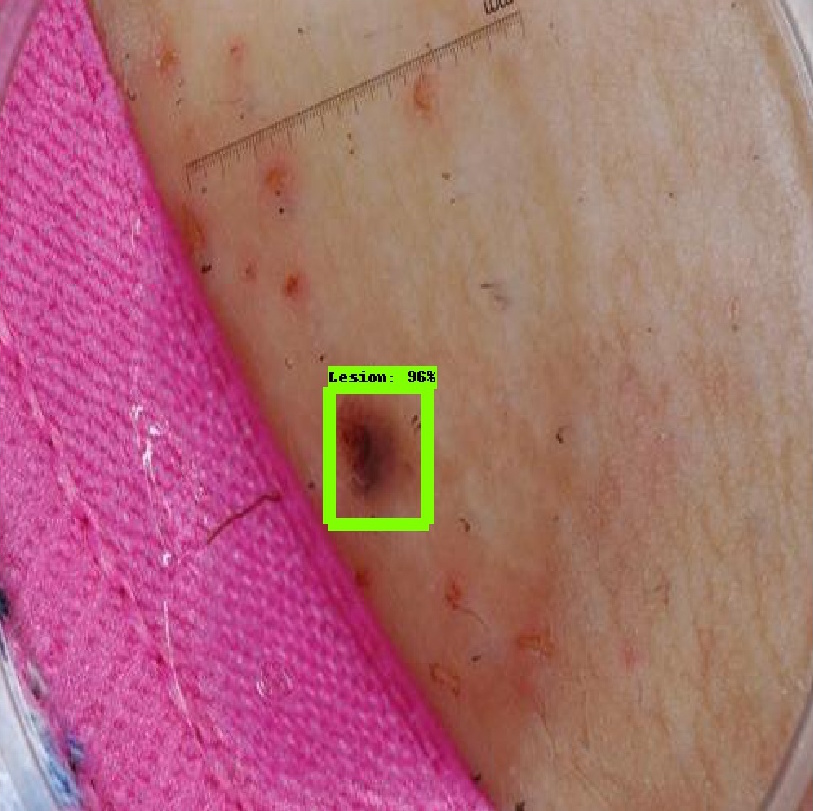} &
		\includegraphics[width=2.5cm,height=2.5cm]{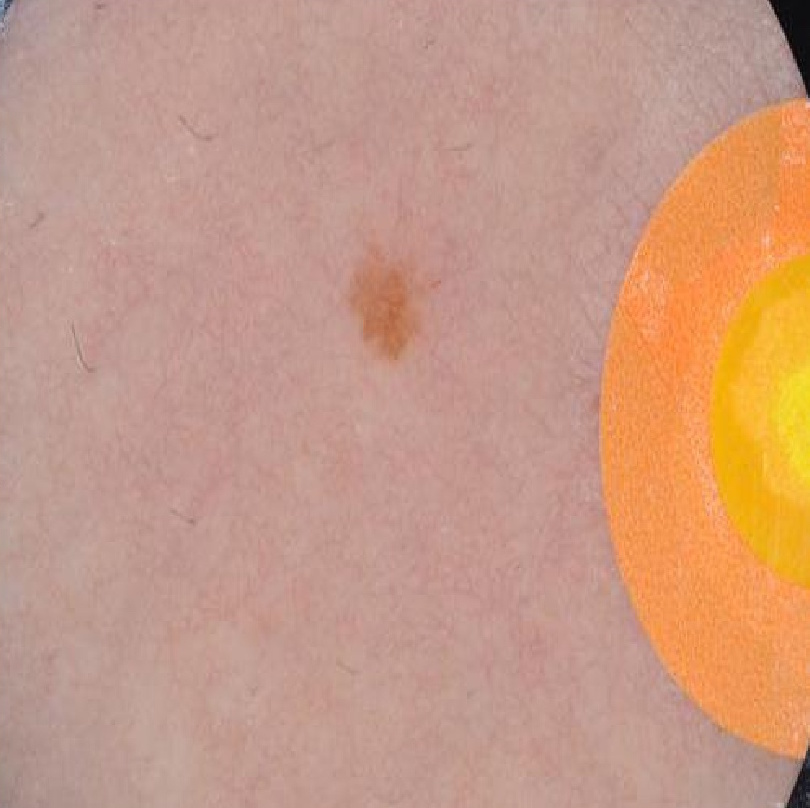}&
		\includegraphics[width=2.5cm,height=2.5cm]{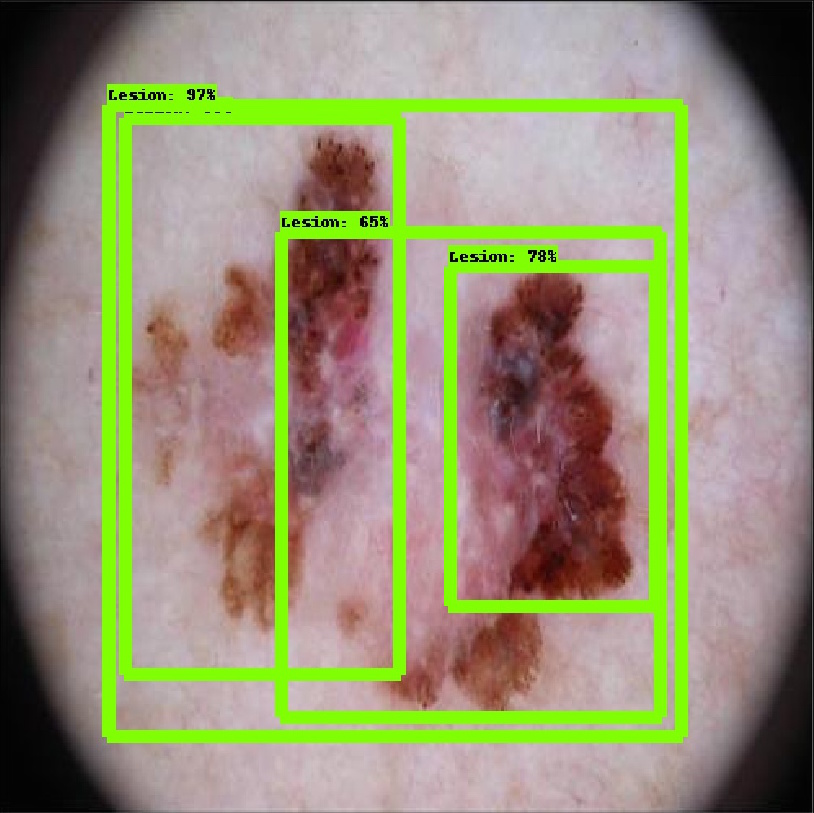}&    
		\\
		(a) Detection  & (b) No-Detection& (c) Multi-Detection &
		\\
	\end{tabular}     
	
	\caption{Examples of inference Produced by Faster R-CNN Inception-V2: (a) \textit{TP}: Detection with \textit{IoU} of 98.2\%; (b) \textit{FN}: No Detection where the model failed to detect the skin lesion; and (c) Multi-Detection: The bounding box with the highest confidence was selected as the detected ROI. The remaining bounding boxes were counted as \textit{FP}.}
	\label{fig:resultsVisual22}
\end{figure}




\section{Experiment and Results} 

We used the 2017 ISIC Challenge dataset to train all the networks on a GPU machine with the following specifications: (1) hardware: CPU - Intel i7-6700@ 4.00 GHz; GPU - NVIDIA TITAN X 12 GB; RAM - 32 GB DDR5; and (2) software: TensorFlow

We trained SSD Inception-V2 for 100 epochs with a batch size of 24; in addition to having an \texttt{RMS\_Prop} optimizer, with a learning rate of 0.004, and a decay factor of 0.95. For transfer learning,  we used a pre-trained model that was trained on the MS-COCO dataset, which consists of more than 80,000 images of 90  classes  \cite{huang2016speed, goyal2017fully}. To train Faster R-CNN Inception-V2, we used a batch size of 2, and a \texttt{Momentum} optimizer with a manual step learning rate, and an initial rate as 0.0002, with a step-down rate of 10\% after every 33 epochs. In Fig. \ref{fig:resultsVisual22},  we show examples of inference produced by the skin detection models. The majority of \textit{FN} cases had very subtle features or similar skin tone or were covered with lots of hair.  For multi-detection cases, we used the bounding box with the highest confidence to determine the final results. In the case of equal confidence scores, we selected the bounding box with the largest area. For each test image, we generated a single bounding box as the detected ROI.
We received state-of-the-art \textit{JSI} Score of 77.15 on the testing set which is better than \textit{JSI} Score achieved by the competition winner (76.5) \cite{codella2017skin}. Similar to the way that we generated ground truth, the circumscribed rectangle bounding boxes are generated from the resulted segmentation mask.   

In Table \ref{tab:tradFeats2w2d}, we indicate the performance of each trained models on ISIC-2017 testing dataset. Overall, Faster R-CNN Inception-V2 performed best in regards to the ROI detection, with a \textit{precision} of 0.945 (94.5\%), \textit{recall} of 0.943 (94.3\%) for \textit{IoU} \textgreater~0.5. It performed well throughout the performance indices used for \textit{IoU} \textgreater~0.75. The DeepLabV3+ scored first position in \textit{mean IoU(0.5)} performance index due to low score of 0.887 in both \textit{precision(0.5)} and \textit{recall(0.5) }. Comparing the performances of SSD Inception-V2 and DeepLabV3+, DeepLabV3+ outperformed SSD Inception-V2 in all of the performance measures except \textit{precision(0.5)} for which SSD Inception-V2 (91.8\%) performed better than DeepLabV3+ (88.7\%)

\begin{figure*}[ht]
	\centering
	\begin{tabular}{ccc}
		\includegraphics[width=5cm,height=4cm]{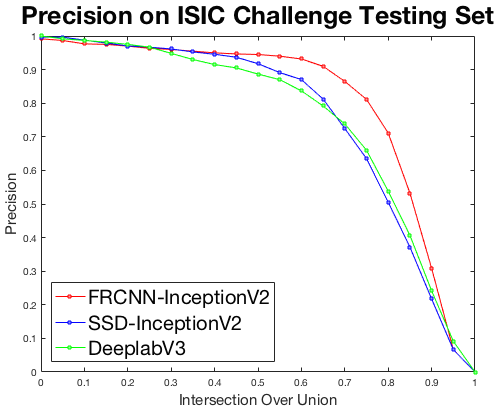} & \includegraphics[width=5cm,height=4cm]{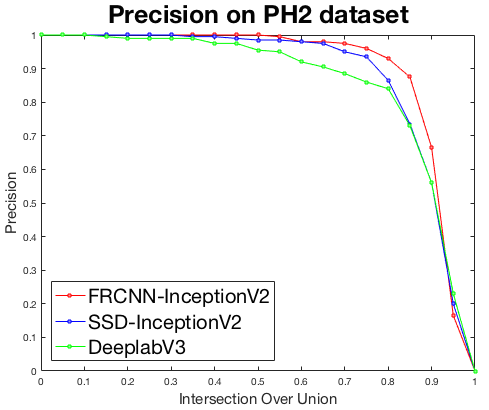}&
		\includegraphics[width=5cm,height=4cm]{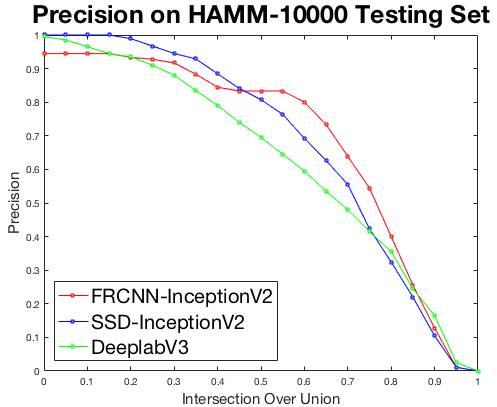}\\
		\includegraphics[width=5cm,height=4cm]{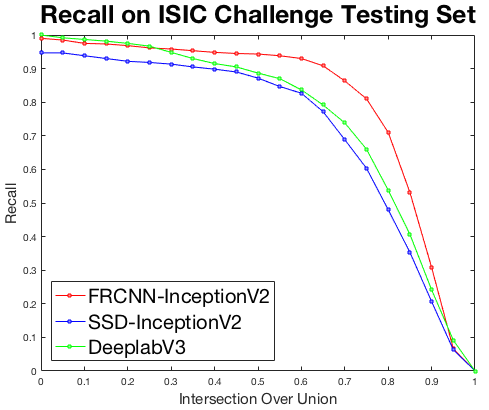}&
		\includegraphics[width=5cm,height=4cm]{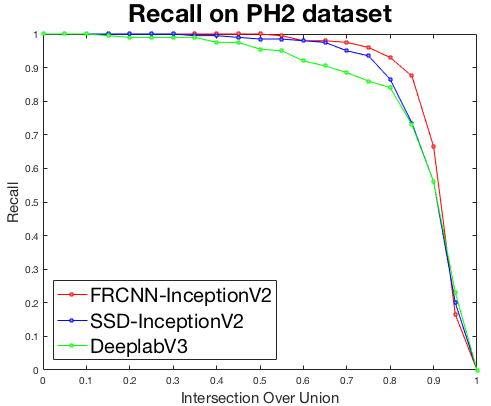} & \includegraphics[width=5cm,height=4cm]{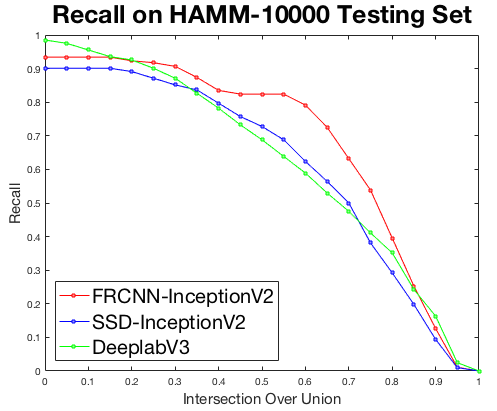} \\
		\includegraphics[width=5cm,height=4cm]{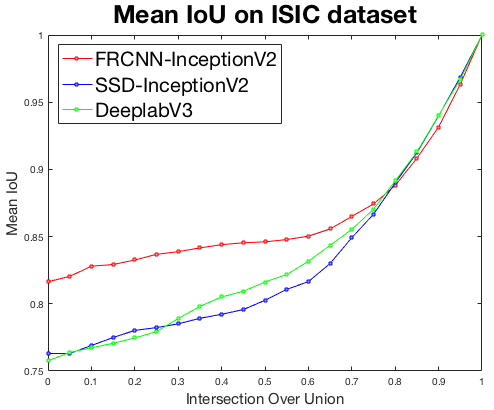} & \includegraphics[width=5cm,height=4cm]{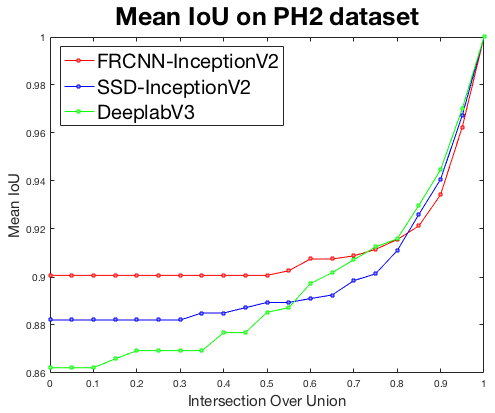}&
		\includegraphics[width=5cm,height=4cm]{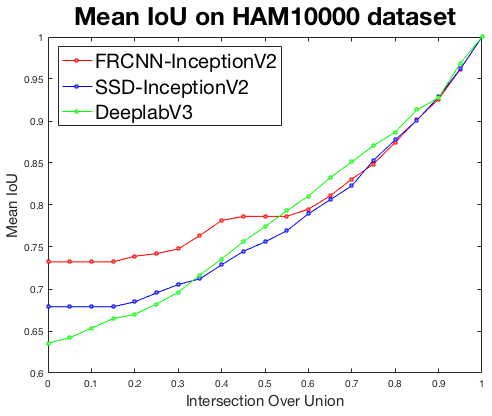} \\
		
	\end{tabular}     
	
	\caption{Performance measure curves of all proposed methods on skin lesion datasets}
	\label{fig:resultsVisual22p}
\end{figure*}

We further investigated the performance of these models on PH2 and a subset of HAM10000. Faster R-CNN Inception-V2 performed better than the other models on these completely unseen datasets as shown in Table \ref{tab:tradFeats2w2d}. Faster R-CNN Inception-V2 performed well on PH2 dataset in terms of \textit{precision(0.5)} and \textit{recall(0.5)}. Moreover, SSD Inception-V2 performance was better in most of the performance measures than DeepLabV3+ on these datasets. We report the curves of performance measures of all proposed methods for the skin lesion detection task in Fig. \ref{fig:resultsVisual22p}. For a subset of the HAM10000 dataset, these skin lesion categories were not available in the training set (ISIC-2017); the ROI detection algorithms still achieved reasonably high accuracy, which proves the generalization of these networks for ROI detection on unseen classes of skin lesions. 

\begin{table*}[]
	\centering
	\large\addtolength{\tabcolsep}{3pt}
	\caption{Comparison of performance for different detection and segmentation methods for ROI detection on ISIC-2017, PH2 and HAM10000 testing set where PR refers Precision.}
	\renewcommand{\arraystretch}{2}
	\label{tab:tradFeats2w2d}
	\scalebox{0.7}{
		\begin{tabular}{cccccccc}
			\hline
			Dataset & Method   & \textit{PR(0.5)} & \textit{Recall(0.5)} & \textit{Mean IoU(0.5)} & \textit{PR(0.75)} & \textit{Recall(0.75)} & \textit{Mean IoU(0.75)}\\ \hline \hline

			\multirow{3}{*}{ISIC 2017} & FRCNN Inception-V2 &  \textbf{0.945}&\textbf{0.943}&0.811 & \textbf{0.810}&\textbf{0.863}&\textbf{0.874}\\ 
			
			&SSD Inception-V2 &  0.918&0.872&0.803 &  0.603 &0.688&0.866\\    
			
			&DeepLabV3+ &  0.887&0.887&\textbf{0.816} &  0.660 &0.738&0.870\\ \hline   
			
			\multirow{3}{*}{PH2} & FRCNN Inception-V2 &  \textbf{1.000}&\textbf{1.000}&\textbf{0.901} & \textbf{0.960}&\textbf{0.960}&\textbf{0.912}\\ 
			
			&SSD Inception-V2 &  0.985&0.985&0.889 &  0.935 &0.935&0.901\\    
			
			&DeepLabV3+ &  0.955&0.955&0.885 &  0.860 &0.860&0.912\\ \hline  
			
			\multirow{3}{*}{HAM10000}& FRCNN Inception-V2 &  \textbf{0.833}&\textbf{0.824}&\textbf{0.786} &  \textbf{0.544}&\textbf{0.539}&0.849\\ 
			
			&SSD Inception-V2 &  0.808&0.728&0.756 &  0.423&0.381 &0.853\\    
			
			&DeepLabV3+ &  0.695&0.688&0.774 &  0.415 &0.411&\textbf{0.870}\\ \hline   
	\end{tabular}}
\end{table*}

\begin{figure}[!t]
	\centering
	\begin{tabular}{cc}
		\includegraphics[width=4cm,height=3cm]{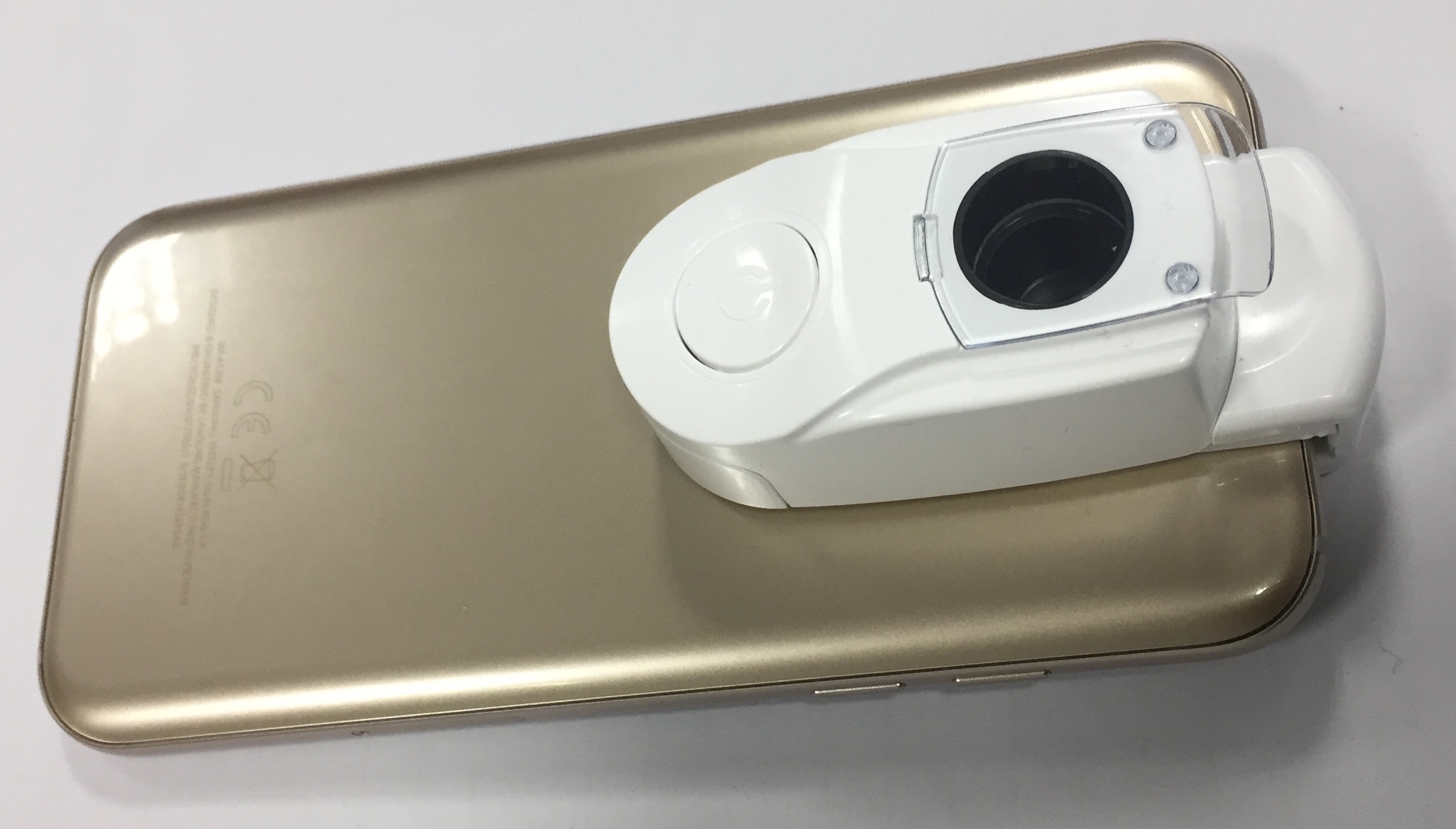} 
		&
		\includegraphics[width=4cm,height=3cm]{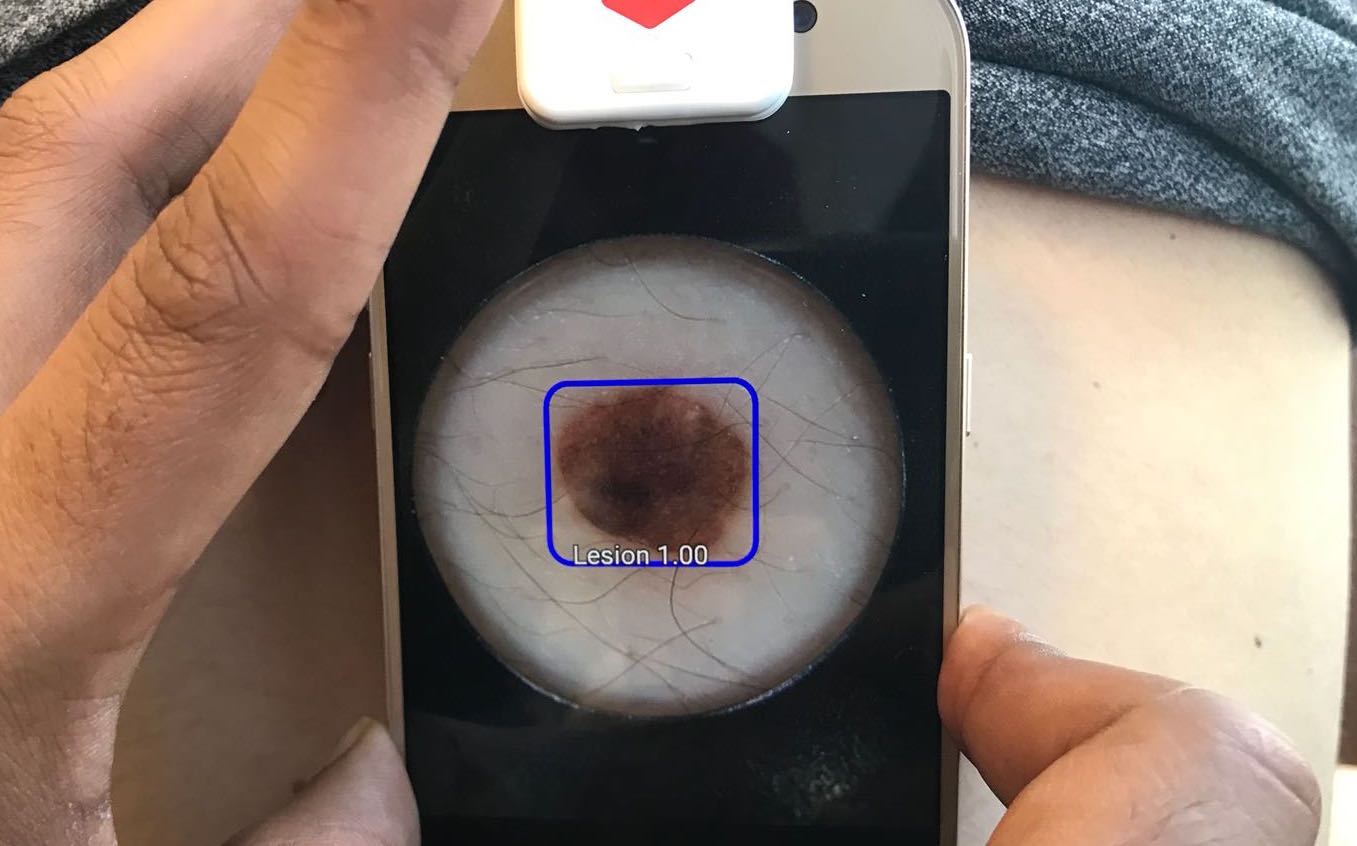}   \\ 
		(a) &
		(b) \\
		
	\end{tabular}     
	
	\caption{Illustration of our real-time ROI detection application on skin lesions: (a) an Android smart-phone with a MoleScope; and (b) Real-time inference with confidence level of 100\%.}
	\label{fig:resultsVisual2222}
\end{figure}  

\begin{figure}[!t]
	\centering
	\begin{tabular}{cc}
		\includegraphics[width=4cm,height=4cm]{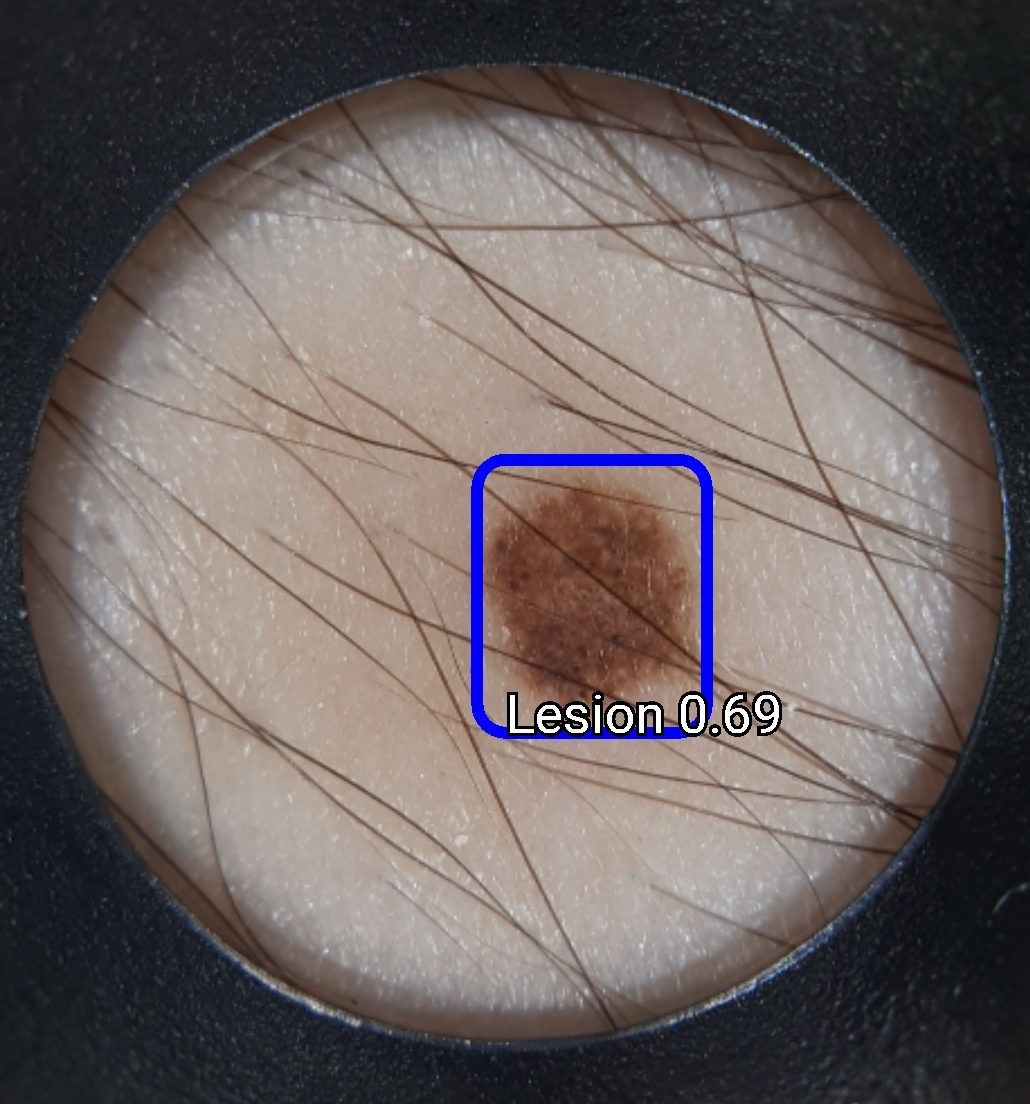} 
		& \includegraphics[width=4cm,height=4cm]{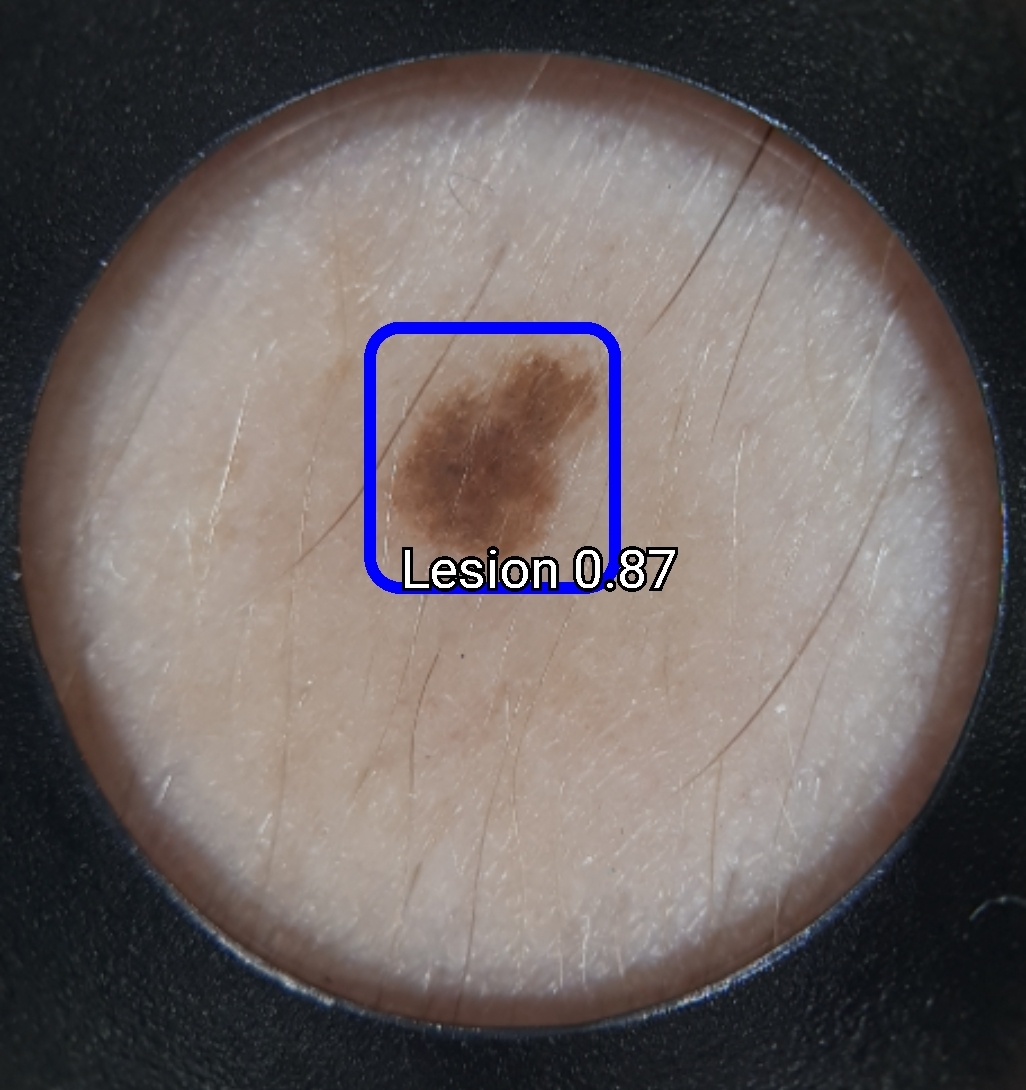} \\
		\includegraphics[width=4cm,height=4cm]{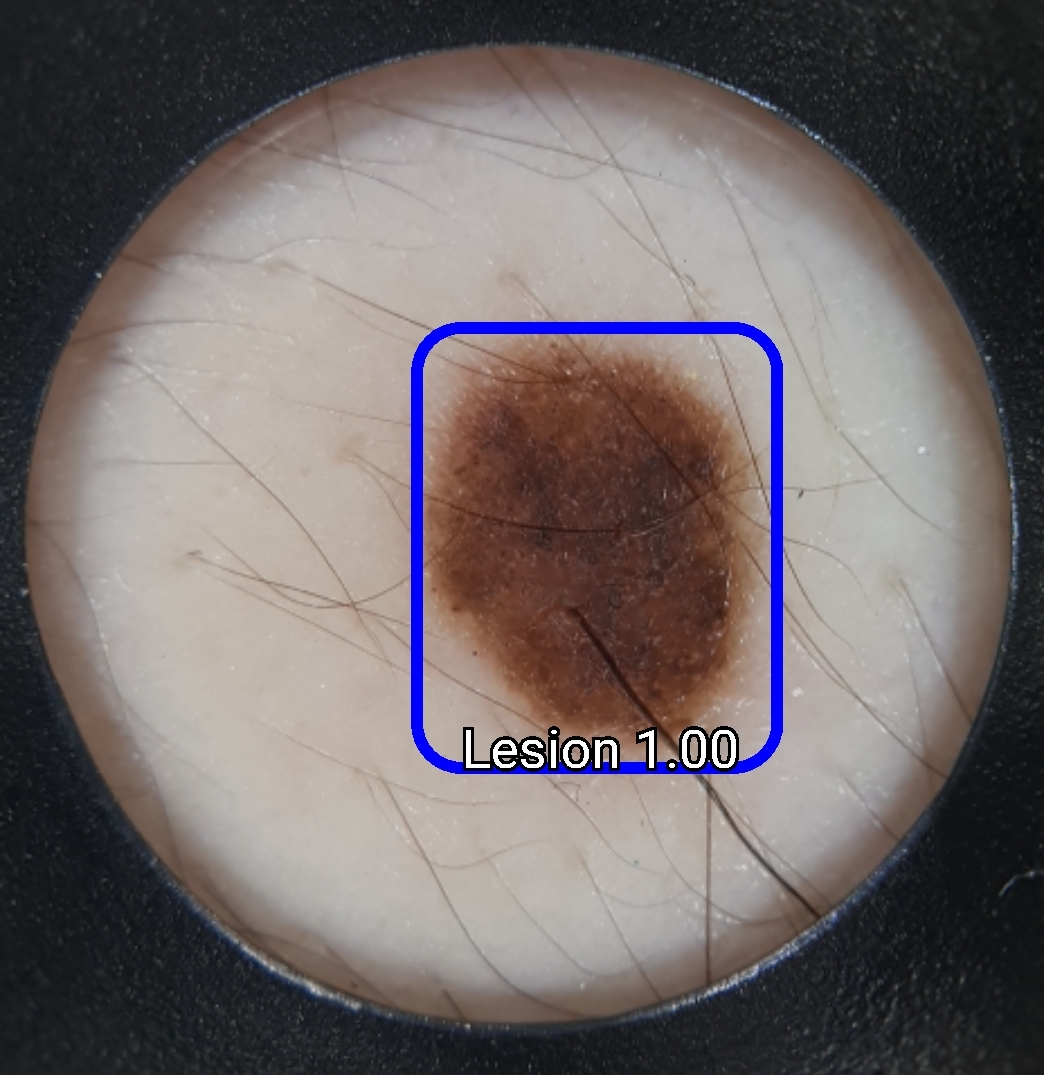} 
		& \includegraphics[width=4cm,height=4cm]{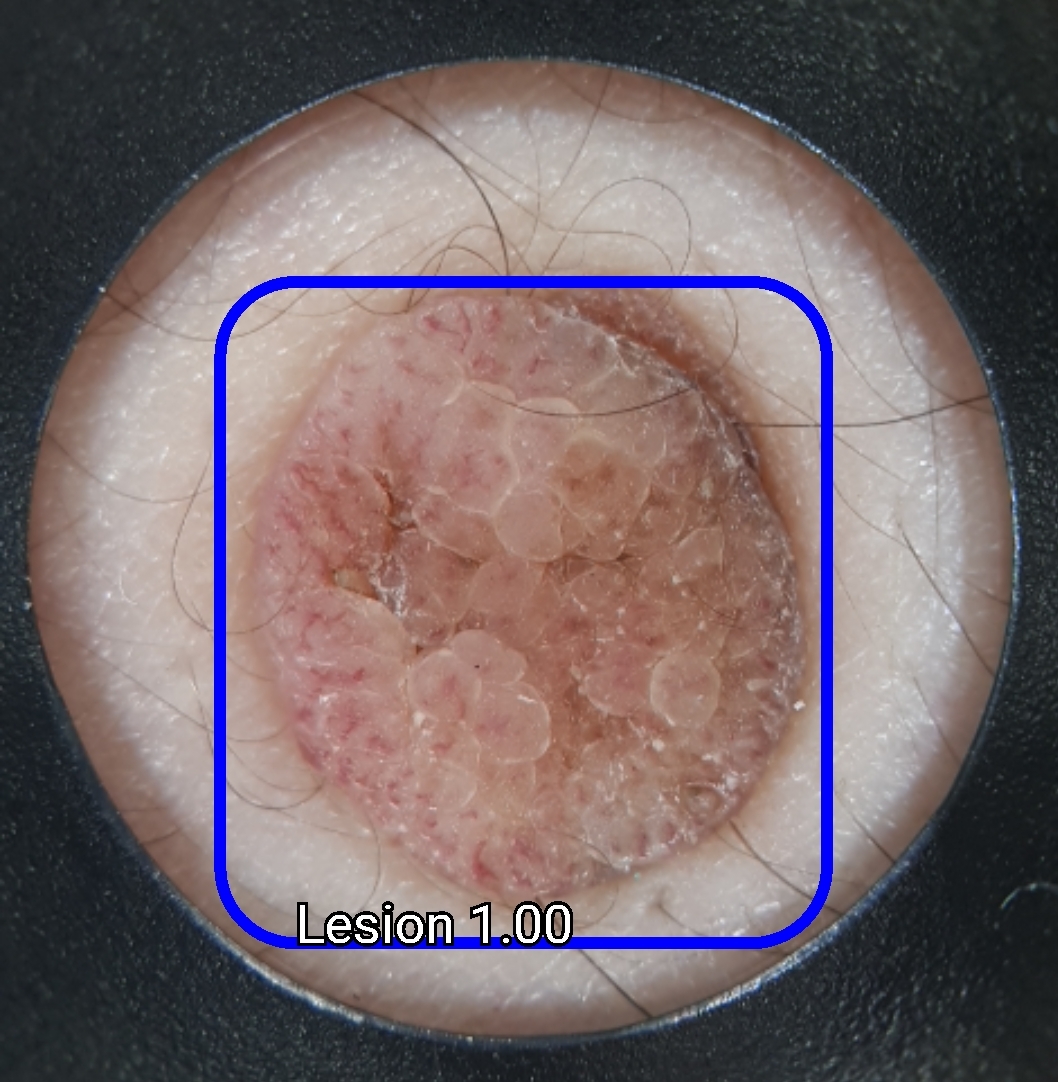} \\

	\end{tabular}     {\tiny }
	
	\caption{Additional examples of lesion detection on smart-phone application}
	\label{fig:resultsVisual22s3}
\end{figure}

\begin{table*}[]
	\centering
	\small\addtolength{\tabcolsep}{2pt}
	\caption{Comparison of latency of Deep Learning detection models}
	\renewcommand{\arraystretch}{2.1}
	\label{tab:tradFeats2}
	\scalebox{0.7}{
		\begin{tabular}{lcc}
			\hline
			Model Name                & Size of Model(MB)  & Average Speed(ms)  \\ \hline \hline
			SSD Inception-V2                & 47.6       &  38    \\
			Faster R-CNN Inception-V2              & 58.6      & 57  \\ \hline
	\end{tabular}}
\end{table*}

\section{Smart-phone Application for Lesion Detection}
For real-time inference on mobile devices, it is important to consider lightweight models (low latency) in terms of the size of the model and inference speed. Although the performance of Faster R-CNN (three-stage architecture) was superior to the SSD algorithm (two-stage architecture) in all datasets, the SSD algorithm is more suitable for mobile deployment due to its low latency. We deployed SSD Inception-V2 for real-time ROI detection on a smartphone using Android Studio and TensorFlow mobile libraries.  Since these models were trained on dermoscopic images, we used a MoleScope attachment for a Samsung A5 smartphone that provides a high-resolution, detailed view of the moles through magnification and specialized lighting (Fig. \ref{fig:resultsVisual2222}). We tested 30 different types of skin lesions on 12 different people with this smartphone application. This application was able to provide accurate real-time detection of skin lesions.  It can be a useful feature for capturing dermoscopic image data using different camera devices.
In Table \ref{tab:tradFeats2}, the size and speed (at inference) of our trained detection algorithms are reported. Additional examples of skin lesion detection are shown in Fig. \ref{fig:resultsVisual22s3}.


\begin{figure}[!t]
	\centering
	\small
	\begin{tabular}{ccc}
		\includegraphics[width=4cm,height=4cm]{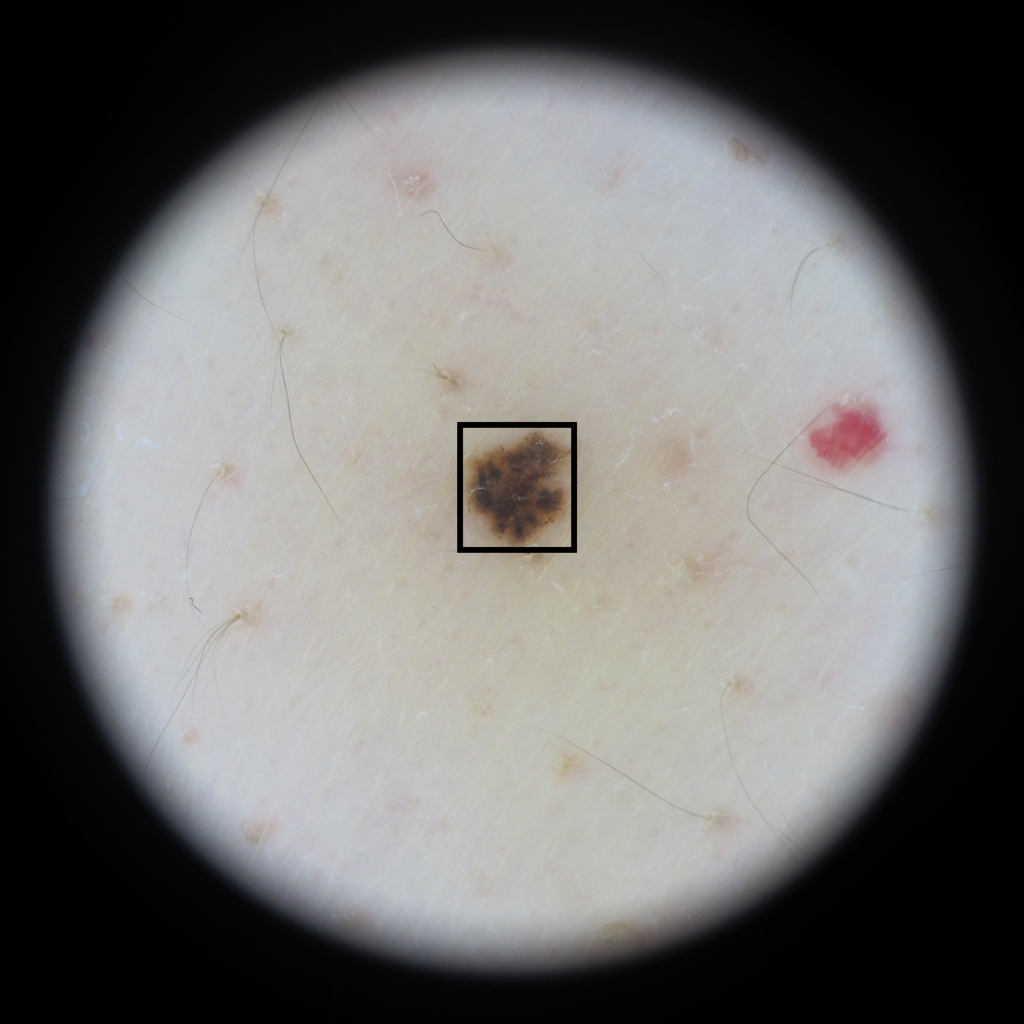} &
		\includegraphics[width=4cm,height=4cm]{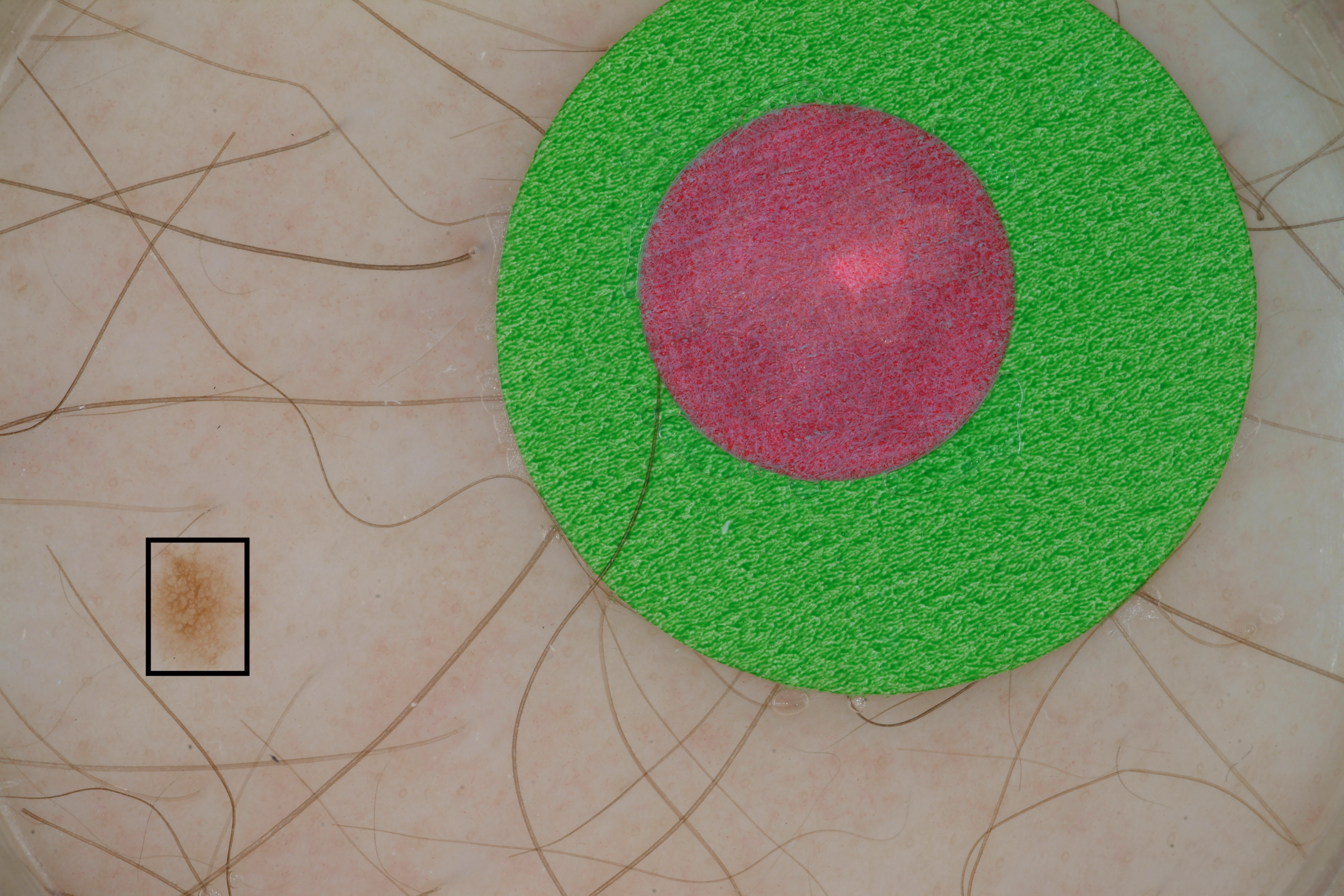}\\
		(a) dark corner & (b) color chart and hair \\
		\includegraphics[width=4cm,height=4cm]{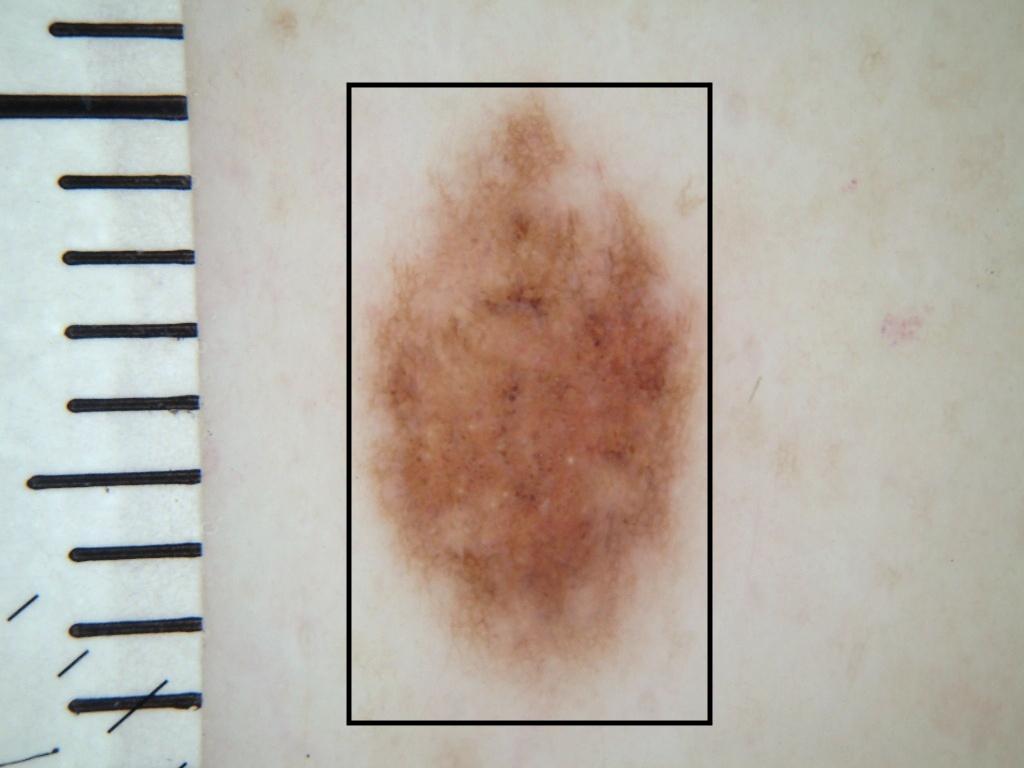}& 
		\includegraphics[width=4cm,height=4cm]{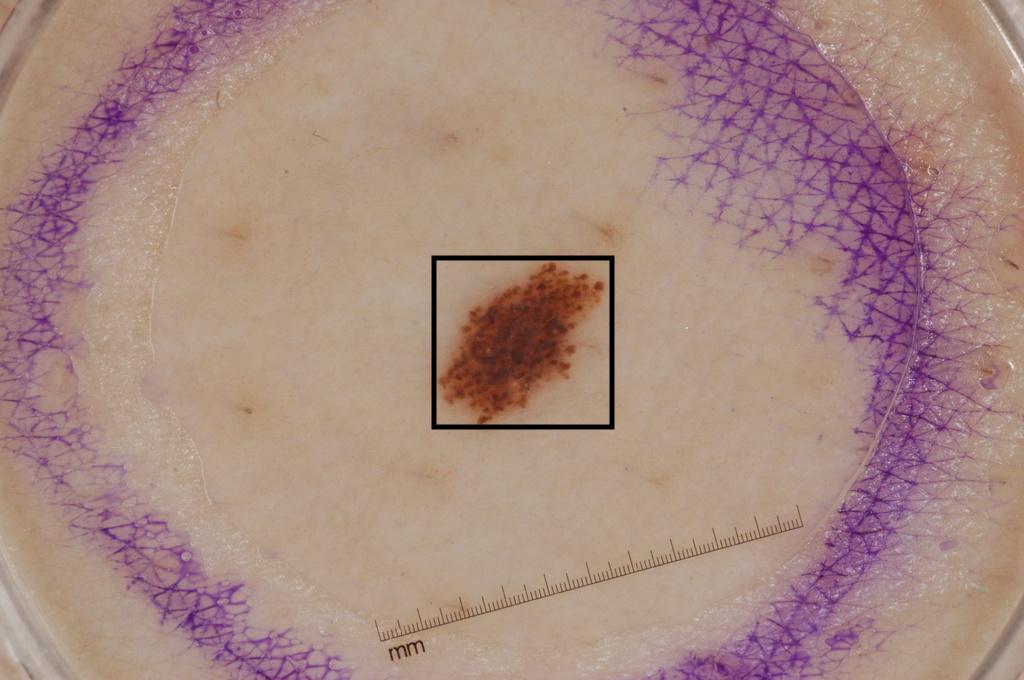} \\
		(c) ruler marker & (d) ink and ruler marker &
		
		\\
	\end{tabular}     
	\caption{Common artifacts in dermoscopic skin lesion images    that may confuse deep learning algorithms \cite{codella2018skin}}
	\label{fig:Artifacts}
\end{figure}

\begin{figure*}[!t]
	\centering
	\small
	\begin{tabular}{ccccc}
		\includegraphics[width=2.85cm,height=2.85cm]{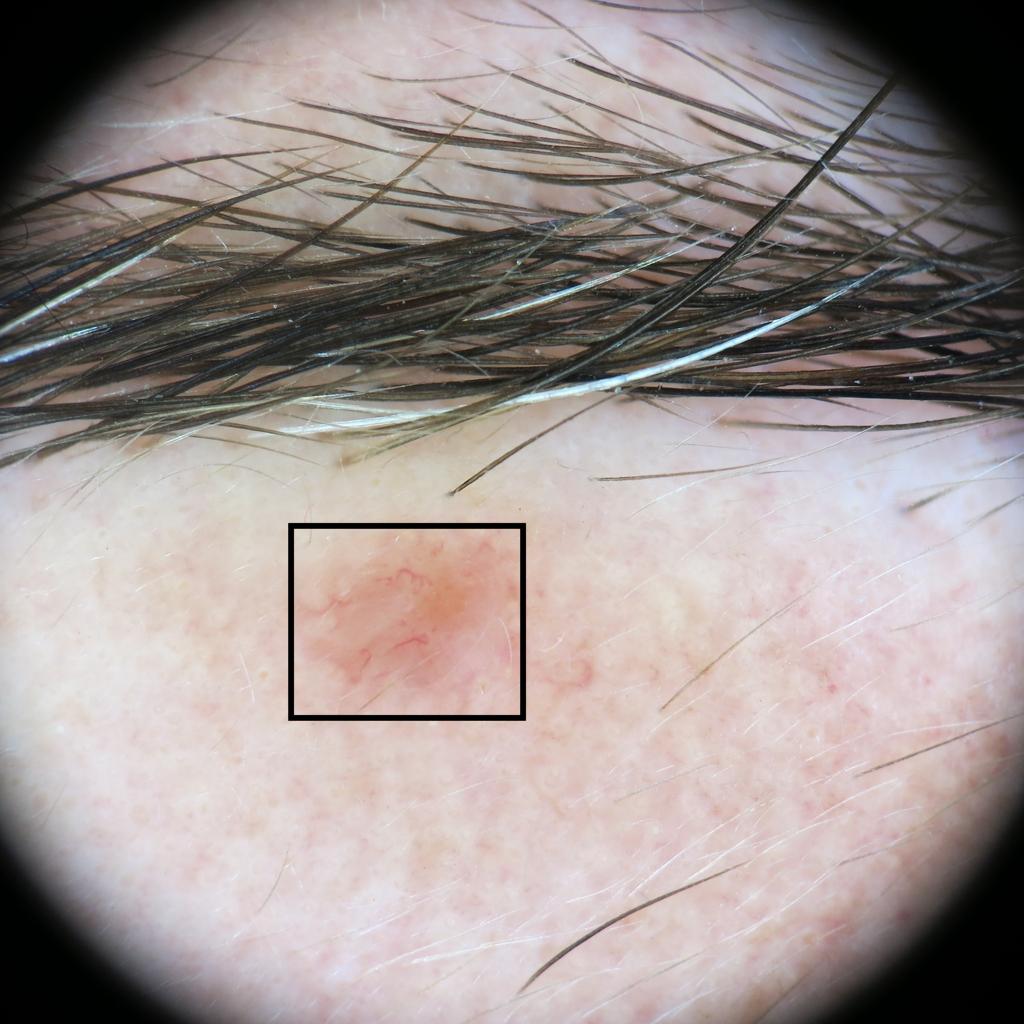} &
		\includegraphics[width=2.85cm,height=2.85cm]{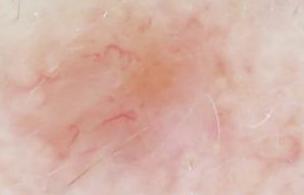}&
		\includegraphics[width=2.85cm,height=2.85cm]{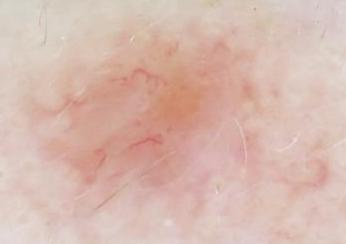}& 
		\includegraphics[width=2.85cm,height=2.85cm]{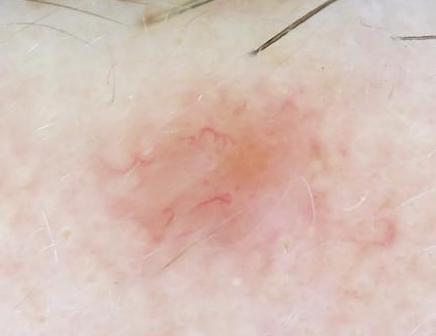}&    
		\\
		(a) Dark Conner & (b) 1st MAG & (c) 2nd MAG & (d) 3rd MAG &
		\\and Hair&&&
		\\
		\includegraphics[width=2.85cm,height=2.85cm]{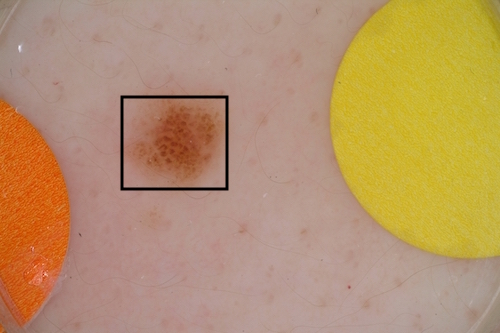} &
		\includegraphics[width=2.85cm,height=2.85cm]{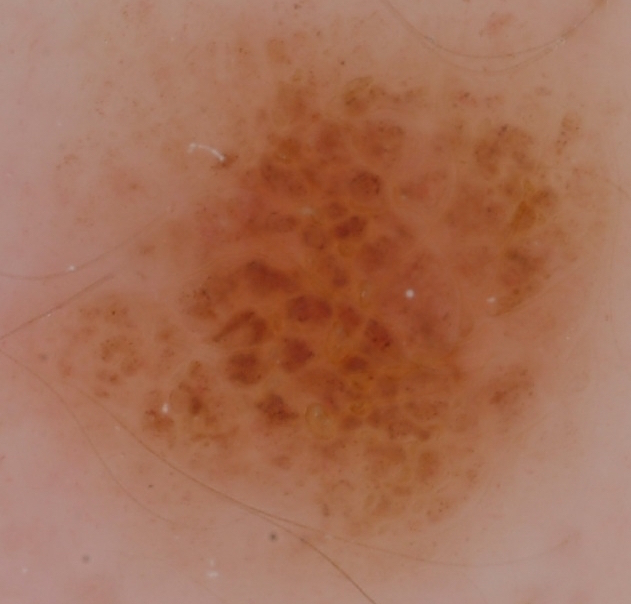}&
		\includegraphics[width=2.85cm,height=2.85cm]{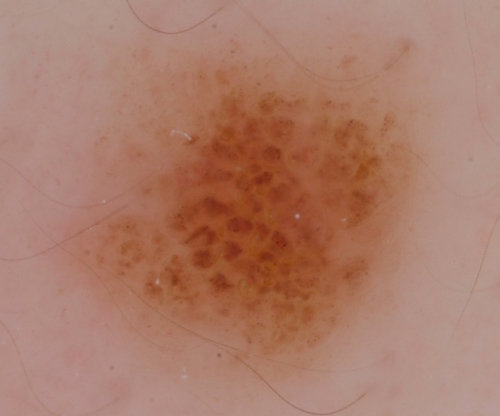}& 
		\includegraphics[width=2.85cm,height=2.85cm]{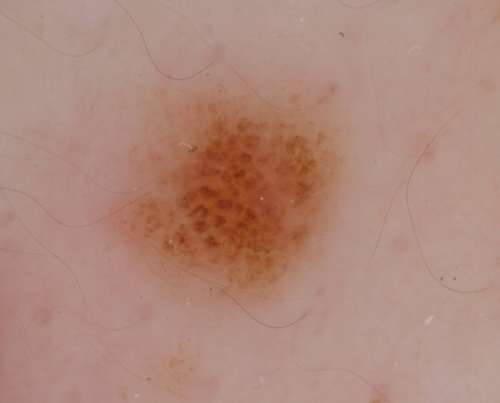}&    
		\\
		(a) Color Chart & (b) 1st MAG & (c) 2nd MAG & (d) 3rd MAG &
		\\
		
		\includegraphics[width=2.85cm,height=2.85cm]{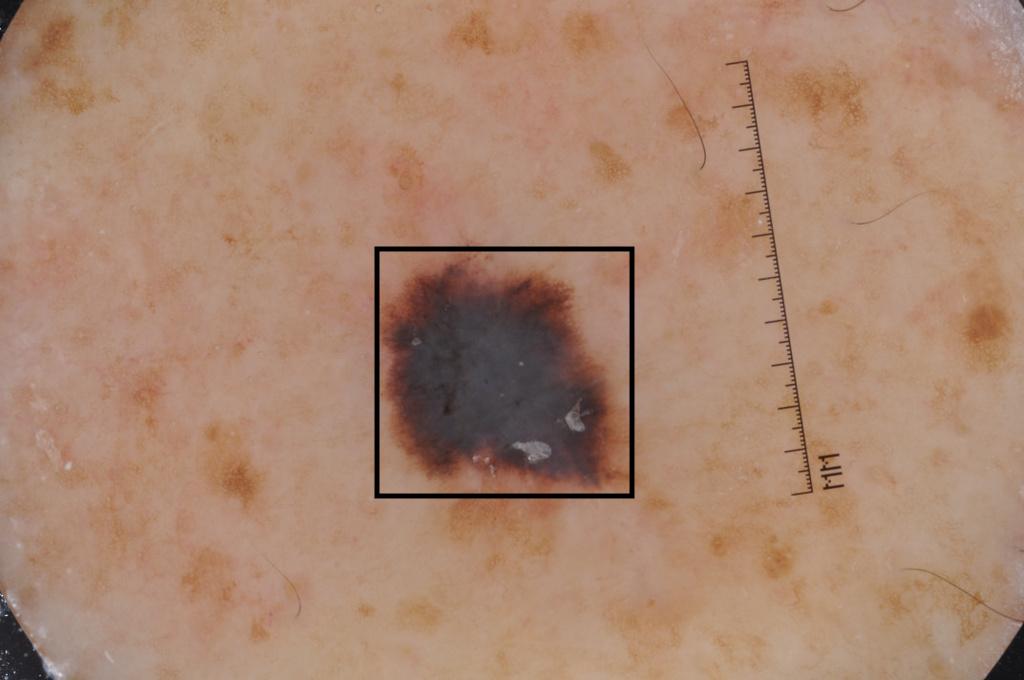} &
		\includegraphics[width=2.85cm,height=2.85cm]{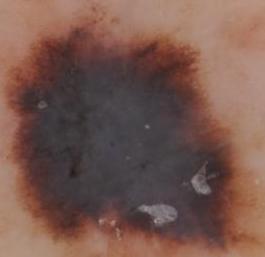}&
		\includegraphics[width=2.85cm,height=2.85cm]{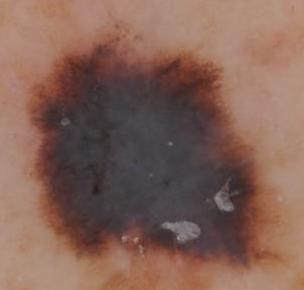}& 
		\includegraphics[width=2.85cm,height=2.85cm]{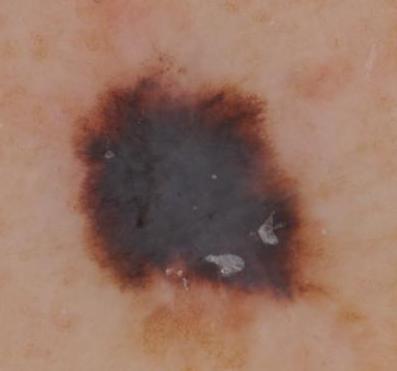}&    
		\\
		(a) Ruler Marker & (b) 1st MAG & (c) 2nd MAG & (d) 3rd MAG &
		\\
		\includegraphics[width=2.85cm,height=2.85cm]{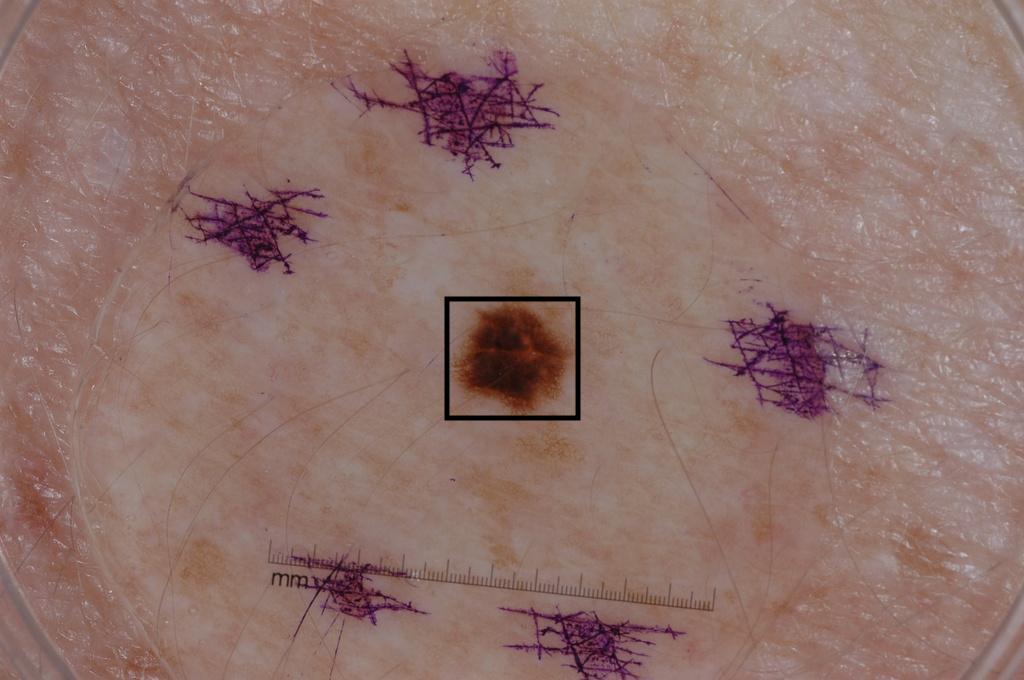} &
		\includegraphics[width=2.85cm,height=2.85cm]{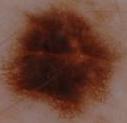}&
		\includegraphics[width=2.85cm,height=2.85cm]{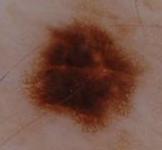}& 
		\includegraphics[width=2.85cm,height=2.85cm]{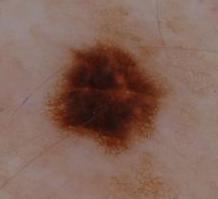}&    
		\\
		(a) Ink Marker & (b) 1st MAG & (c) 2nd MAG & (d) 3rd MAG &
		\\

	\end{tabular}     
	
	\caption{Natural data-augmentation produced from the original image with different magnifications to highlight ROI of skin lesions and remove unnecassary artifacts. The abbreviation ``MAG" refers to magnification}
	\label{fig:resultsVisual223}
\end{figure*}

\begin{figure*}
	\centering
	\includegraphics[scale=0.62]{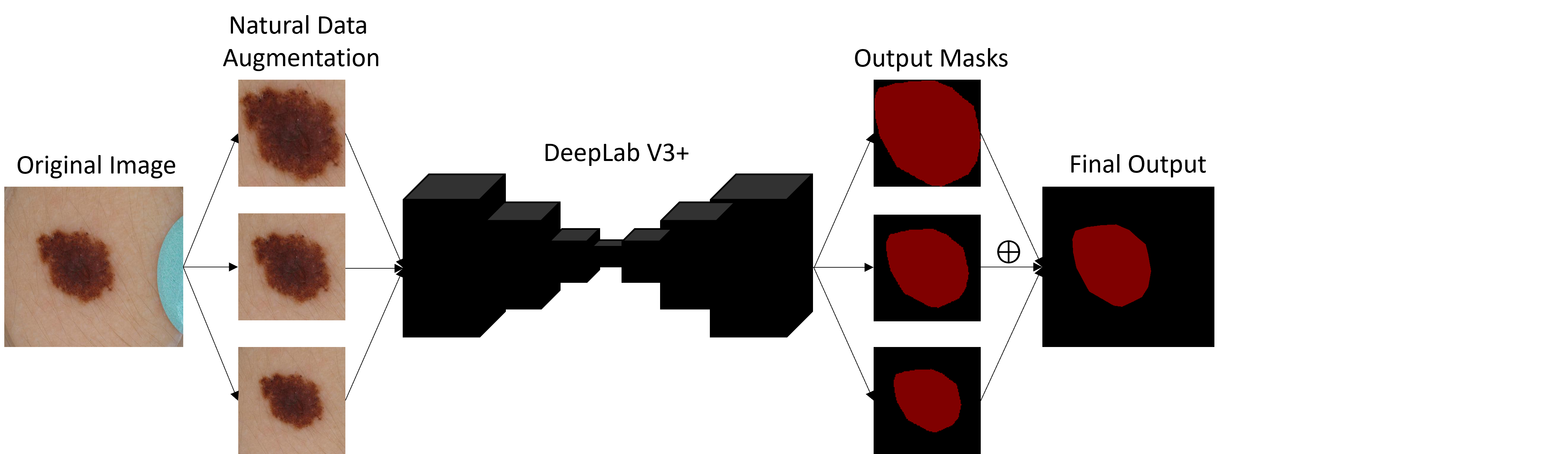}
	\caption{Segmentation results produced using ROI detection as pre-processing and DeeplabV3+ using natural data-augmentation on ISIC-2017 testing set.}
	\label{fig:NatAuggg}
\end{figure*}

\section{Natural Data-Augmentation}

In publicly available dermoscopic datasets, the images are captured with different devices and techniques all around the world. There could be many unnecessary artifacts other than skin lesions in dermoscopic images, and that may confuse computer-aided diagnosis and detection (CAD) systems (Fig. \ref{fig:Artifacts}). In contrast, our proposed Natural data-augmentation method performs data augmentation to highlight important ROI on skin lesion images at different magnification levels as well as removes these unnecessary artifacts (Fig. \ref{fig:resultsVisual223}). In dermoscopic datasets, the size of images varies between 540$\times$722 and 4499$\times$6748, as these images are captured with a professional camera. On the other hand, most of the deep learning algorithms use smaller image sizes, such as 224$\times$224. For natural data-augmentation, multiple augmented copies of the same skin lesion with different magnifications are generated without redundant data or any loss of quality.

The number of natural data-augmentation methods with detection depends upon the input image size for the algorithm and the ratio between the size of the ROI and the size of the image. As the ROI of skin lesions is highlighted with different magnifications, we can achieve other data-augmentation techniques, such as rotation with different angles using our proposed method (Supplementary Fig. 6). On the other hand, without our proposed method, there is a chance that popular traditional data-augmentation techniques, such as random crop and translation, could miss important ROIs on images of skin lesions, especially when the size of the skin lesion is minimal compared to the overall size of the dermoscopic image (Supplementary Fig.  7).

\subsection{Skin Segmentation Results using Natural Data-Augmentation}

In this section, we checked the effectiveness of the proposed natural data- augmentation technique as a pre-processing step for the segmentation of skin lesions and compared it with state-of-the-art segmentation algorithms. Standard hyperparameters were used to train the state-of-the-art deep learning algorithms for skin segmentation. For example, for Fully Convolutional Networks (FCN), we trained DeepLabV3+ for skin lesion segmentation on the 2017 ISIC Challenge segmentation dataset images, using natural data-augmentation and traditional data-augmentation the author used for the segmentation of the Pascal-VOC dataset. For traditional data-augmentation, we used rotation, translation, random crop ($>$0.75), horizontal, and vertical flip on the fly during training. For natural data-augmentation, we used 24,356 training and 1,792 validation images, respectively, from the original 2,000 training and 150 validation images. The inference for the test images was produced (Fig. \ref{fig:NatAuggg}).

In Table \ref{my-label2st}, we reported the performance measures of the state-of-the-art segmentation algorithms and our proposed DeeplabV3+ method using natural data-augmentation on the 2017 ISIC Challenge segmentation dataset. The results showed that the proposed method segmented the skin lesions with a \textit{Jaccard index} of 82.3\% for the 2017 ISIC Challenge test dataset. In comparison to CDNN, U-Net, FCN, SegNet, DeepLabV3+, and DeepLabV3+ (traditional data-augmentation), the proposed method (DeepLabV3+ with natural data-augmentation) outperformed them by 5.8\%, 6.1\%, 8.3\%, 12.7\%, 5.2\%, and 3.5\% for the \textit{Jaccard index}, respectively.

\begin{table*}[]
	\centering
	\addtolength{\tabcolsep}{3pt}
	\renewcommand{\arraystretch}{2}
	\caption{Performance evaluation of state-of-the-art segmentation algorithms and DeeplabV3+ with natural data-augmentation trained and tested on ISIC-2017 segmentation dataset}
	
	\scalebox{0.7}{
		\begin{tabular}{cccccc}
			\hline
			Methods      & Accuracy & Dice &Jaccard &Sensitivity&Specificity  \\ \hline \hline
			First: Yading Yuan (CDNN Model)  \cite{yuan2017automatic}   &0.934 &0.849    & 0.765 & 0.825 & 0.975\\ 
			Second: Matt Berseth (U-Net) \cite{ronneberger2015u} &0.932 &0.847    & 0.762 & 0.820 & \textbf{0.978}\\  
			FCN \cite{long2015fully} &0.927 &0.828    & 0.736 & 0.811 & 0.967\\ 
			SegNet \cite{badrinarayanan2015segnet} &0.918 &0.821    & 0.696 & 0.801 & 0.954\\
			Mask R-CNN \cite{he2017mask} & 0.935 & 0.856    & 0.774 & 0.848 & 0.960\\
			DeepLabV3+ \cite{chen2018deeplab} &0.936 &0.851    & 0.771 & 0.843 & 0.972\\ 
			DeepLabV3+ (Trad Data-augmentation) &0.939 &0.866    & 0.788 & 0.887 & 0.965 \\
			DeepLabV3+ (Natural Data-augmentation)   &\textbf{0.947} & \textbf{0.886}    &  \textbf{0.823} & \textbf{0.898} & 0.964 \\ \hline
	\end{tabular}}
	\label{my-label2st}
\end{table*}


\section{Conclusion and Future Work}
In this work, we proposed the use of deep learning methods to detect and localize skin lesions. We trained and tested our methods on the 2017 ISIC Challenge dataset. We evaluated the performance of these models on completely unseen the PH2 dataset and a subset of the HAM10000 dataset with high accuracy. The Faster R-CNN Inception-V2 method outperformed the state-of-the-art segmentation model for ROI detection on all testing sets. ROI detection and data-augmentation are two important aspects of medical image analysis. Further, we designed a real-time smartphone application for ROI detection using a lightweight model (SSD Inception-V2) on an Android device attached to a MoleScope. 

We demonstrated the potential of this ROI detection work to propose a natural data-augmentation method that produces the augmentation of skin lesions at different magnifications and angles. It is an alternative approach to traditional data augmentation methods commonly used for machine learning applications. With natural data-augmentation as a pre-processing step, the performance of state-of-the-art deep learning algorithms for skin segmentation was significantly improved. Natural data-augmentation was also effectively used as a pre-processing step for deep learning methods used for the recognition of ischemia and infection in diabetic foot ulcers as shown in foot images \cite{goyal2020recognition}. In the future, this work can be extended to train deep learning models to combine both classification and detection of ground truths for multi-class skin lesion detection to diagnose different types of skin cancer. We also plan to perform a comparative analysis of natural data-augmentation and traditional data-augmentation for machine learning algorithms on different datasets as future work .

\vspace{6pt} 



\section*{Author Contributions}
The manuscript was written by M.G. under the supervision of M.H.Y and S.H. The modeling and software process was executed by M.G. S.H. and M.H.Y. helped in the analysis and review.

\section*{Funding}
This research was supported in part by National Institutes of Health grants R01LM012837 and R01CA249758.


\section*{Conflicts of Interest}
No  conflicts  of  interest,  financial  or  otherwise,  are  declared  by  the  authors.

\bibliographystyle{IEEEtran}
\bibliography{skin}

\begin{IEEEbiography}[{\includegraphics[width=1in,height=1.25in,clip]{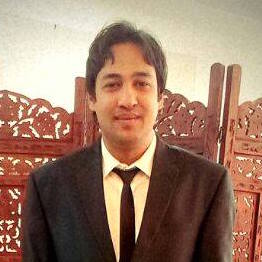}}]{Manu Goyal} 
	has a research background in artificial intelligence with a Ph.D. degree in artificial neural networks/ deep learning/ medical imaging entitled “Novel Computerised Techniques for Recognition and Analysis of Diabetic Foot Ulcers” from Manchester Metropolitan University. He is currently a Postdoctoral Research Associate with the department of radiology, UTSouthwestern Medical Center, Dallas. His research expertise is in medical imaging analysis, computer vision, deep learning, wireless sensor networks, and the Internet of Things. He serves as a reviewer of the IEEE Transactions/Journals (Image Processing, Access, Medical Imaging, Biomedical Health, and Informatics)
\end{IEEEbiography}

\begin{IEEEbiography}[{\includegraphics[width=1in,height=1.25in,clip,keepaspectratio]{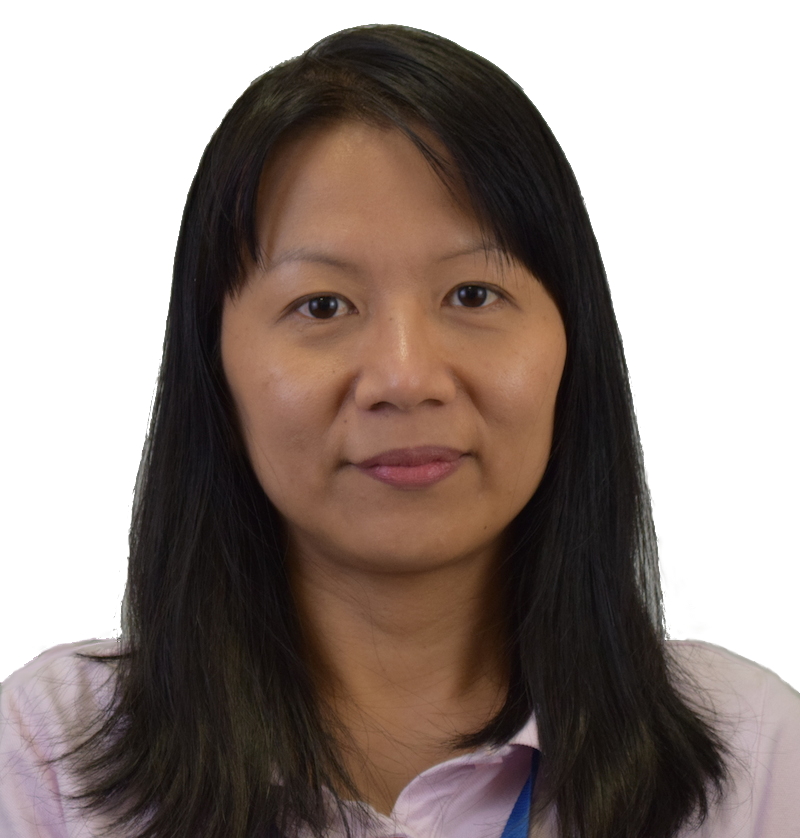}}]{Moi Hoon Yap} is Professor in Computer Vision at the Manchester Metropolitan University and a Royal Society Industry Fellow (2016-2018) with Image Metrics Ltd. She received her Ph.D. in Computer Science from Loughborough University in 2009. After her Ph.D., she worked as Postdoctoral Research Assistant (April 09 - Oct 11) in the Centre	for Visual Computing at the University of Bradford. She serves as an Associate Editor for Journal	of Open Research Software and reviewers for IEEE transactions/journals (Image Processing, Multimedia, Cybernetics, biomedical health, and informatics). Her research expertise is computer vision, applied machine learning and deep learning.
\end{IEEEbiography}

\begin{IEEEbiography}[{\includegraphics[width=1in,height=1.25in,clip,keepaspectratio]{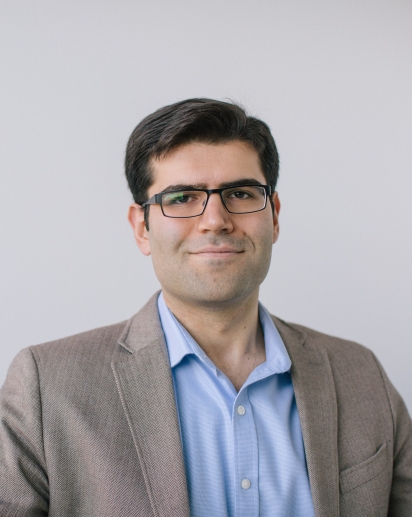}}]{Saeed Hassanpour} is an Associate Professor in the Departments of Biomedical Data Science, Computer Science, and Epidemiology at Dartmouth College. His research is focused on the use of artificial intelligence in healthcare. Dr. Hassanpour’s research laboratory has built novel machine learning and deep learning models for medical image analysis and clinical text mining to improve diagnosis, prognosis, and personalized therapies. Before joining Dartmouth, he worked as a Research Engineer at Microsoft. Dr. Hassanpour received his Ph.D. in Electrical Engineering with a minor in Biomedical Informatics from Stanford University and a Master of Math in Computer Science from the University of Waterloo in Canada.
\end{IEEEbiography}

\end{document}